\documentclass{article}
\usepackage{spconf,amssymb,amsmath,graphicx}
\usepackage{epsfig}
\usepackage{xspace}
\usepackage{srcltx}
\usepackage{caption}
\usepackage{subcaption}
\usepackage[export]{adjustbox}
\usepackage{multirow,makecell,hhline}
\usepackage{booktabs}
\usepackage{textcomp}
\usepackage{cite}
\usepackage{soul}
\usepackage{url}
\usepackage{hyperref}
\hypersetup{colorlinks=True, urlcolor=black}

\usepackage{setspace}

\usepackage{color}
\usepackage[dvipsnames]{xcolor}

\newcommand{\CMT}[1]{{}}

\title{Efficient Adapter Finetuning for Tail Languages in Streaming Multilingual ASR
}

\name{Junwen Bai, Bo Li, Qiujia Li, Tara N. Sainath, Trevor Strohman}
\address{Google, USA \\
\fontsize{9}{9}\selectfont\ttfamily\upshape
\{junwen,boboli\}@google.com}
\begin{document}
\ninept
\maketitle
\begin{abstract}
The end-to-end ASR model is often desired in the streaming multilingual scenario since it is easier to deploy and can benefit from pre-trained speech models such as powerful foundation models. Meanwhile, the heterogeneous nature and imbalanced data abundance of different languages may cause performance degradation, leading to asynchronous peak performance for different languages during training, especially on tail ones. Sometimes even the data itself may become unavailable as a result of the enhanced privacy protection. Existing work tend to significantly increase the model size or learn language-specific decoders to accommodate each language separately. In this study, we explore simple yet effective Language-Dependent Adapter (LDA) finetuning under a cascaded Conformer transducer framework enhanced by teacher pseudo-labeling for tail languages in the streaming multilingual ASR. The adapter only accounts for 0.4\% of the full model per language. It is plugged into the frozen foundation model and is the only trainable module during the finetuning process with noisy student training. The final model merges the adapter parameters from different checkpoints for different languages. The model performance is validated on a challenging multilingual dictation dataset, which includes 39 tail languages across \textit{Latin}, \textit{Greek}, \textit{Arabic}, etc. Our proposed method brings 12.2\% word error rate reduction on average and up to 37.5\% on a single locale. Furthermore, we show that our parameter-efficient LDA can match the quality of the full model finetuning, thus greatly alleviating the asynchronous peak performance issue. 

\end{abstract}
\noindent
\section{Introduction}
\label{sec:intro}

End-to-end multilingual automatic speech recognition (MASR) is an active research topic where a single speech model is trained to recognize multiple languages \cite{pratap2020massively,li2021scaling,babu2021xls}. A single MASR system is often more cost-efficient to deploy compared to a large number of monolingual models \cite{li2022massively}. However, training an end-to-end MASR system is not trivial. Different languages have distinct vocabularies and data abundances. A single model may not have enough capacity to accommodate all languages/locales \cite{zhang2023google}. Recent efforts focus on increasing the model size. For instance, USM \cite{zhang2023google} reaches model size of 2B parameters and MMS \cite{pratap2023scaling} has over 1B parameters. These large-scale models can often serve as foundation models for non-streaming scenarios \cite{li2023efficient,openai2023gpt4}. It is desirable to utilize their large capacity with minimal changes for different languages, especially the tail ones.

Nevertheless, while these efforts have incorporated more languages and shown improved qualities, it is still not clear how the performance varies across all languages. Many existing papers select and report the checkpoint with the lowest average word error rate (WER) across all the tested languages. Such WER is often contributed by only a portion of all the languages while sacrificing other under-fitting or over-fitting locales. For example, in a recent MASR model \cite{bai2022joint}, the best WER on \textit{Portuguese} is achieved at 100K steps, while \textit{Polish} is already over-fit and \textit{Dutch} is still under-fit at this checkpoint. As a trade-off, the checkpoint at 100K is selected since its average WER is the lowest. It is hard to ensure the model can reach optimal performance for all languages at the same checkpoint. We call this problem the asynchronous peak performance issue. To this end, prior work such as MMS \cite{pratap2023scaling} increases the model size and inserts language-specific adapters and heads during finetuning for each language. However, models with billions of parameters often have latency concerns and are not suitable for applications that require streaming processing \cite{sainath2022improving}. Also, each language-specific adapter is trained for 2K steps uniformly, ignoring the heterogeneity across languages. Language-specific heads add up to 2\% extra parameters per language, which results in an extra burden for deployment. On the other hand, ASR transducer models with cascaded encoders have shown to be a better fit to process the streaming speech \cite{li2021better} by achieving good trade-off between latency and quality. A recent study \cite{mavandadi2023truly} proposed a language-agnostic 1st pass and a language-aware 2nd pass under the cascaded architecture with promising results. However, this work requires a non-causal encoder and decoder for every language respectively, resulting in a total model size of 0.6B.

\begin{figure}[t]
	\centering
	\includegraphics[width=0.27\textwidth]{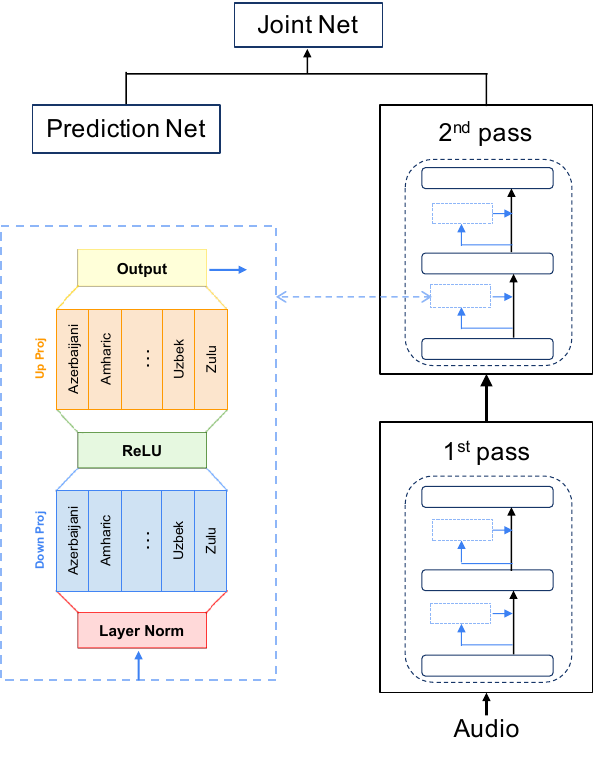}
    \caption{An overview of LDA in a Conformer model with cascaded encoders. LDAs are inserted between two consecutive Conformer layers for both 1st and 2nd passes. Each LDA module contains a stack of language-dependent parameters.
    }
	\label{fig:lda}
\end{figure}

To address the aforementioned issues, in this work, we introduce Language-Dependent Adapters (LDAs), which can achieve significant improvements for streaming MASR models with cascaded encoders. Based on a frozen pre-trained foundation model, we add and train the LDAs for tail languages. Since the backbone model is frozen, we can collect and gather adapter weights from different steps. During the LDA finetuning, an off-the-shelf well-trained foundation model \cite{li2022massively,li2022language} is adopted as the backbone, which is a stack of Conformer layers \cite{gulati2020conformer}. LDAs are inserted as a residual pass at the end of each Conformer layer. The input to the next Conformer layer is the combination of the output from the previous layer and the LDA output. We support a batch of mixed languages during finetuning by adding language ID to each utterance \cite{waters2019leveraging}. Language IDs are converted to one-hot vectors to select the corresponding language-dependent weights in each LDA module. Therefore, each utterance would only update the adapter for the corresponding language without interfering with other languages. During training, the foundation model is frozen. For each language, the related LDA weights only take up 0.4\% of the full model size, and we do not add extra projection heads for separate languages, maintaining a total size of 0.2B. After the training process is completed, we select the checkpoints with the best performances for each language. Since they all share the same foundation model, we simply extract LDA weights from those checkpoints and merge them to compose the final LDA module. Therefore, by assembling the checkpoints of peak performances into a single one, we can keep most merits of a single end-to-end model like deployability and low latency while improving the quality. 
Meanwhile, such an asynchronous strategy eases the pressure of carefully balancing the utterances from different languages during training. 
Our LDA follows the prior adapter designs \cite{houlsby2019parameter,hou2021exploiting,pratap2023scaling,li2023modular} but is adapted neatly to fit the MASR domain. Figure \ref{fig:lda} illustrates an overview of LDA.

During the optimization of LDA, we also incorporate noisy student training (NST) \cite{xie2020self,park2020improved} to utilize the unlabeled data. In our task, NST starts with a non-streaming ASR teacher model and iteratively generates pseudo-labels for unlabeled utterances, as well as a series of student ASR models, with the help of a fixed language model (LM) trained separately. The final student during this iterative process would be our streaming LDA model.

Our MASR model with LDAs is validated on a challenging dictation benchmark, specifically on tail languages including \textit{Latin}, \textit{Greek}, \textit{Arabic}, \textit{Cyrillic}, etc. On average, our model can bring 12.2\% relative improvement on WER among all the 39 languages. If we narrow down the scope to the 27 languages with higher priority in the online traffic, the improvement can rise to 17\%. On individual tail languages like \textit{Hungarian} and \textit{Serbian}, the WER reduction can even reach 35\%, which implies that without comprehensive weight updates or architecture re-design, there is still considerable room to improve in the streaming MASR model. Moreover, we demonstrate that LDA finetuning can achieve performance on par with the full model finetuning where all the parameters including the foundation model are refined for individual languages, which further proves the value of adapter finetuning for streaming MASR: (1) a small number of learnable parameters, (2) flexibility of merging different checkpoints, and (3) promising WER reductions.




\section{Related Work}
\label{sec:related}

Our framework builds on many prior studies. In \cite{hou2021meta}, Meta-Adapter uses meta-learning to implicitly transfer the learned knowledge from source languages to one unseen target language by updating the adapters through gradients of languages sampled from training data. SimAdapter \cite{hou2021exploiting} further adopts an attention mechanism to learn the similarity between source and target languages, forcing the general knowledge transferred from the pre-trained model to the tested language. This work focuses on the monolingual transfer and yet the improvement is not clear when the multilingual task scales up. \cite{winata2020adapt} attempts to learn a common adapter to distill the language-agnostic information, and to facilitate language-specific updates from a loss perspective. The concern for these methods emerges from the overhead during deployment, and the difficulty in learning the common adapter soars when the number of languages increases. 
Unlike many previous methods that focus on monolingual finetuning, our LDA method targets 39 languages and supports mixed language batch finetuning, which further facilitates the adapter tuning and NST when the number of languages grows. Besides, most of these prior works explore the non-streaming scenario, while our primary focus is the streaming case. We further compare the adapter finetuning with the full model finetuning to validate the capacity of the adapter-based methods, which complements many prior studies. This model also extends from the previous paper \cite{kannan2019large} but with more sophisticated Conformer architecture and 4 times more tail languages. The training data amounts of our 39 tail languages are all less than 4\% of those high-resource languages like English. Other efforts towards this direction include utilizing self-supervision and pretraining for low-resource languages \cite{bapna2022mslam}, refined sampling to balance different richness \cite{xiao2021adversarial}, and mixture of experts \cite{sun2023building,hu2023mixture}.

\section{Method}
\label{sec:method}

\begin{figure*}[t!]
	\centering
	\includegraphics[width=0.87\textwidth]{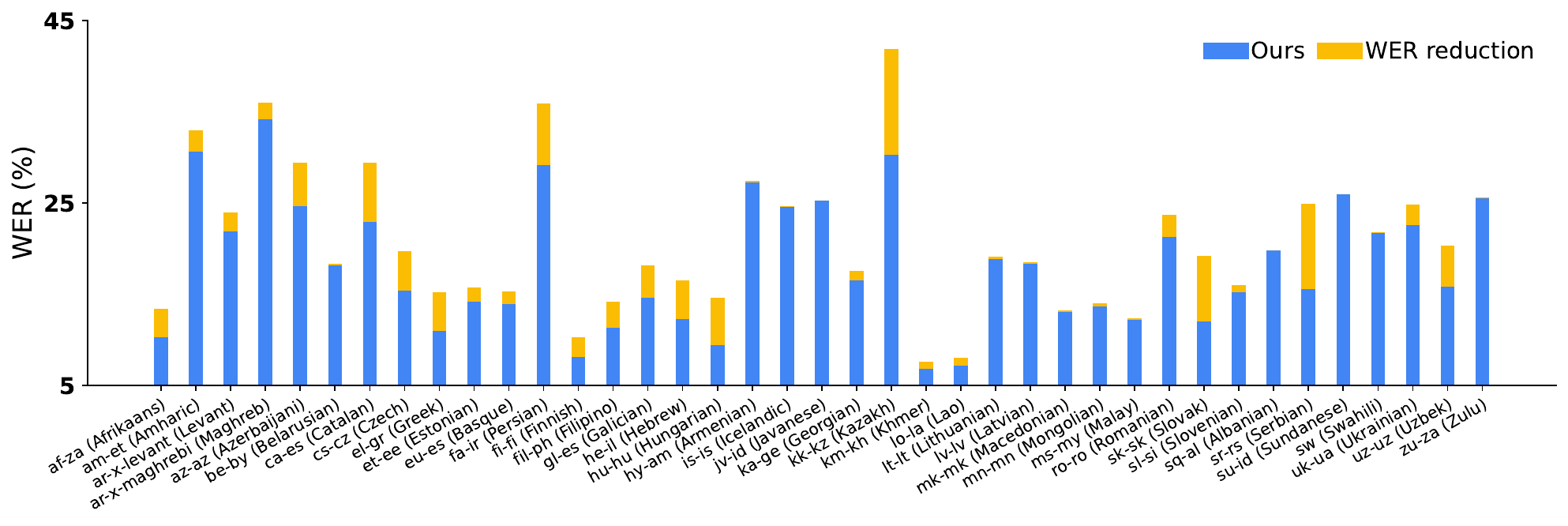}
    \caption{The improvements brought by our method compared to the baseline which is also the existing launched model on the dictation dataset. The blue bars demonstrate the WERs on each language given by our model. The yellow bars highlight the WER reduction outperforming the baseline. The combination of yellow and blue bars denotes the baseline WERs. As shown in the figure, we can achieve significant gains on most languages. On \textit{Slovak}, the gain can reach up to 37.5\%. On average, the improvement on all locales is 12.2\%.
    }
	\label{fig:over_base}
\end{figure*}

Our backbone foundation model follows a cascaded framework \cite{mavandadi2023truly} using a Conformer transducer \cite{gulati2020conformer}, with a causal 1st pass and a non-causal 2nd pass. Unlike \cite{mavandadi2023truly}, both passes are shared across all the languages. The adapters are only inserted to the encoders. The inserted lightweight LDAs contain language-specific parameters during the adapter-only finetuning.

\subsection{Adapter Module}
\label{subsec:adapter}

The LDA architecture resembles the existing adapter modules \cite{houlsby2019parameter}. The output from the previous layer $x_{i-1}\in \mathbb R^{B\times T\times d}$ is passed to the inserted adapter module, where $B$ is the batch size, $T$ is the utterance length and $d$ is the feature dimension. Note that under the streaming scenario, the right context for each time step is masked out during training. Each utterance is associated with a language ID, so we also have $l_{i-1}\in \mathbb N^{B\times K}$ where $K$ denotes the total number of languages (in our case, $K=39$). The down projection matrix $D\in \mathbb R^{Kd\times h}$ maps $x_{i-1}$ to a lower-dimensional space ($h \ll d$) where $h$ is the hidden dimension and it is typically much smaller than $d$. Similarly, the up projection matrix $U\in \mathbb R^{Kh\times d}$ projects the hidden representations back to the $d$-dim space. $l_{i-1}$ would firstly be used to select the language-dependent weights from $D$ and $U$ for the input $x_{i-1}$, $D_x\in \mathbb R^{B\times d\times h}$ and $U_x\in \mathbb R^{B\times h\times d}$. Then the output is given by
\begin{equation}
x_i'=U_x(\textsc{ReLU}(D_x(\textsc{LN}(x_{i-1}))))+x_{i-1}
\end{equation}
where $\textsc{LN}(\cdot)$ represents the pre-LayerNorm \cite{ba2016layer}, a standard normalization to standardize the inputs of the adapter module, and $\textsc{ReLU}(\cdot)$ is the activation function in the hidden space. $x_i'$ is the input for the next layer. We can also further add bias terms $D_b\in \mathbb R^{K\times h}, U_b\in \mathbb R^{K\times d}$ for the down projection and up projection respectively. During LDA finetuning using the Adam optimizer, only $D, U, D_b, U_b$ are trainable and the backbone foundation model remains frozen. Under this framework, LDA supports batch training with mixed languages without concerning their mutual interference. Besides, one can mix the utterances from the full set or a subset. Peak performance can be achieved from different runs and the language-dependent weights from them can be merged together to compose a single end-to-end model.

While many aforementioned adapter variations like common adapters and balancing loss in training could potentially further promote the quality, we found in practice LDA is a good trade-off among the deployability, extensibility, and quality. New languages can simply be appended to the projection matrices with parameter-efficient finetuning. If we have new training data incoming, only a small portion of the full model needs to be updated, without soliciting extra infrastructure changes. More importantly, as we will show in experiments, such a straightforward scheme can already achieve impressive improvements over the existing models in the streaming dictation task, and the final performance can match the expensive full model finetuning on most of the tested tail languages.

\subsection{Noisy Student Training}

The general idea of NST is to use a teacher model trained with supervised data to transcribe unsupervised data in a progressive manner. As better privacy protection has become a societal consensus \cite{voigt2017eu}, NST plays an increasingly critical role when the amount of supervision reduces. In our task, we adopt a non-streaming model as the initial teacher, trained with supervised data and SpecAugment \cite{park2019specaugment}. Then we fuse the teacher with an off-the-shelf LM. The unlabeled data are transcribed using the fused model and a portion of the transcriptions are selected based on the normalized filtering score \cite{park2020improved}. The selected portion is then mixed with the supervised data for training the student model. We take 4 iterations in total and in the last iteration, the trained student model is the LDA model.
\section{Experiments}
\label{sec:exps}

\begin{figure*}[t!]
	\centering
	\includegraphics[width=0.87\textwidth]{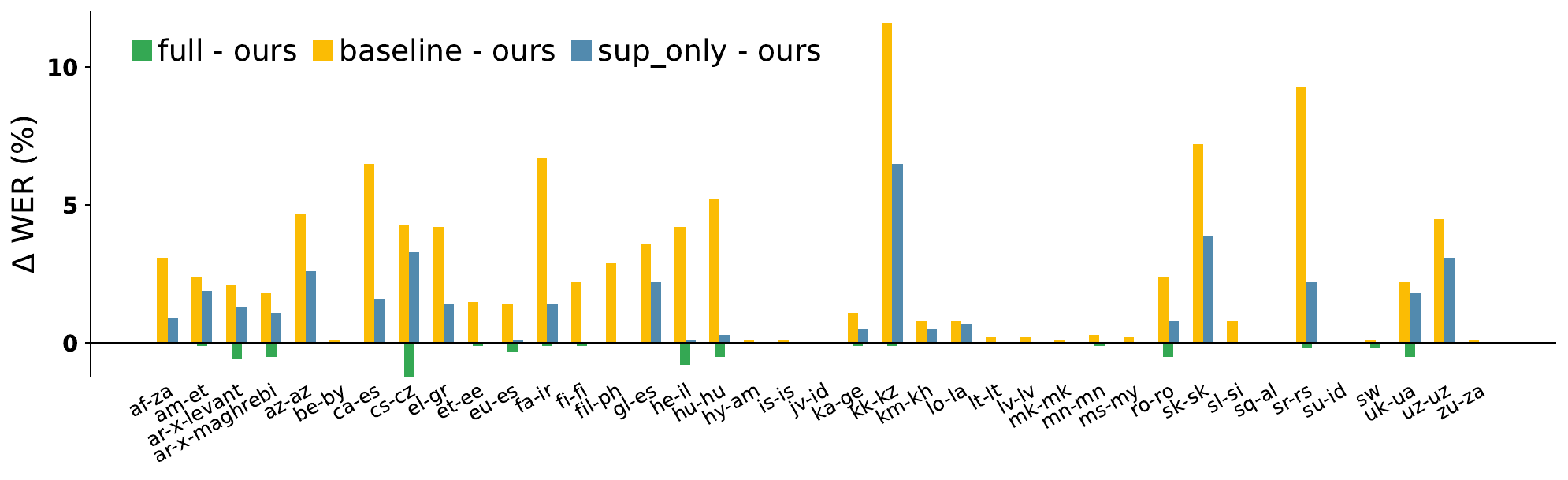}
    \caption{We further compare our model with the full model finetuning. The yellow bars are the same as Fig.~\ref{fig:over_base}. The green bars represent the gap between our model and updating all the parameters for individual locales. As shown in the figure, for most locales, our LDA's performance is on par with the full model finetuning, while ours only updates a small portion of all the parameters. Even on other languages like \textit{Czech, Hebrew}, the yellow bars outweigh the green bars. The blue bars demonstrate our improvements over the baselines trained with supervised data only, proving the contributions from NST.
    }
	\label{fig:full}
\end{figure*}

\subsection{Datasets}

We use a dictation dataset across 39 locales, including \textit{Latin} (\textit{Albanian}, \textit{Icelandic}, \textit{Slovak}), \textit{Arabic} (\textit{Levant}, \textit{Maghrebi}), \textit{Cyrillic} (\textit{Macedonian}, \textit{Kazakh}), \textit{Devanagari} (\textit{Nepali}), etc. These are all tail languages in the online traffic. For instance, \textit{Slovak} utterances only accumulate to 14 hours, which is not even close to 0.01\% of the \textit{English} data.
For all the available supervised data, they are anonymized, and manually transcribed. We have in total 15K hours of audio across 39 languages. The total duration for each language ranges from approximately 14 to 4K hours. 
Besides the human transcriptions, we also incorporate unlabeled data in noisy student training. The amount of unlabeled data is usually much more than the labeled data. As an example, while \textit{Slovak} only has 14-hour human transcriptions, 700-hour audio-only \textit{Slovak} utterances are also available. Similarly, on \textit{Uzbek}, besides the 20-hour human transcriptions, there is also 670-hour audio-only data. Overall, we have 150K unsupervised data and they are effectively utilized using the NST scheme.
  
The test set for each language consists of 12K utterances on average. Test sets are collected separately from the online dictation traffic, and reserved for testing exclusively. The test data are also anonymized and manually transcribed for the evaluation purpose.

\subsection{Architecture}


We use the same core architecture as \cite{li2022language,mavandadi2023truly}. The inputs are 128-dimensional log Mel filterbank computed on 32ms windows with a 10ms hop. 4 contiguous frames are stacked to form a 512-dim input representation with a 30ms frame rate. We use SpecAugment to improve the robustness of our model against noise. In practice, 2 frequency masks with max length 27 and 2 time masks with max length 50 are used. 

We use ten 512-dim Conformer layers in the causal encoder. Each Conformer layer contains causal convolution and left-context attention layers, which strictly excludes future inputs. For each self-attention layer, there are 8 heads and the Convolution kernel size is 15.
The non-causal encoder consists of 7 cascaded Conformer layers. Each Conformer layer is followed by an adapter. Every adapter has a down projection layer and an up projection layer. As described in section \ref{subsec:adapter}, each adapter accommodates 39 languages.

The transducer decoder consists of an embedding prediction network and a joint network. Models use the same vocabulary with 4,096 wordpieces shared by each locale, generated from the pool of transcripts across all languages. The prediction network operates on the previous 2 non-blank model predictions and maps each to a 640-dimensional embedding using a separate 640$\times$4,096 dimensional embedding table. The joint network has 640 dimensions. 


Our model is trained with Tensorflow under the Lingvo framework \cite{shen2019lingvo} on Tensor Processing Units (TPUs) \cite{jouppi2020domain}. The transducer loss \cite{graves2012sequence} is chosen for the model training using the factorization proposed by Hybrid Autoregressive Transducer (HAT) \cite{variani2020hybrid} to allow the incorporation of a language model. FastEmit \cite{yu2021fastemit} is used with a regularization weight of 5e-3.  The batch size is 4,096. Up to 512 TPU cores are used during the optimization using synchronized stochastic gradient descent. Adam optimizer \cite{kingma2014adam} is configured with parameters $\beta_1$=0.9 and $\beta_2$=0.999. Exponential moving average and learning rate schedule with peak learning rate 1.8e-3 follows \cite{vaswani2017attention} to stabilize the weight updates.

\subsection{Results}

We first compare our LDA results with the strong existing end-to-end model \cite{li2022language} in streaming MASR. Compared to \cite{li2022language} (baseline), our model incorporates efficient language-aware adapter modules and scales up 4 times of the validated languages (all on tail ones excluding mainstream ones such as English). In Fig.~\ref{fig:over_base}, LDA achieves improvements on most languages. To highlight the difference between our results and the baseline, we stack the difference on top of our WER numbers (yellow bars). During the LDA finetuning, different locales may obtain peak performance at different checkpoints. However, since the backbone foundation model is frozen, checkpoints only differ in the adapter module. This is a critical consideration when we choose the adapter module for this task. Then we can collect the checkpoints at distinct steps for all languages respectively to synchronize their peak performance and maximize the merits of the end-to-end ASR models. Meanwhile, the light-weight adapters can minimize the risk of over-fitting which hurts the performance. Even if the adapter finetuning might still bring some over-fitting concerns, one can opt to zero out the corresponding weights to skip the adapter module. As a result, we can guarantee that adding adapters can always have a beneficial impact. Fig.~\ref{fig:over_base} demonstrates this point. On 27 out of 39 locales tested, we achieved noticeable improvement by 17\% on average. On \textit{Slovak}, \textit{Serbian}, \textit{Hungarian}, the relative improvements reach 37.5\%, 37.3\%, 35.6\% respectively. On 12 locales, the relative WER improvements are over 20\%. On 20 locales, the relative WER improvements are at least 10\%. The average gain across all languages is over 12\%, which validates the effectiveness of the LDA finetuning. 

We further compare our model with the full model training (also with NST) where all the parameters including the foundation model can be learnable during the training for each language in both passes. Such monolingual finetuning is often adopted in the existing literature \cite{conneau2020unsupervised}. Though such strategies are less favorable in the real-world use since weights often drastically change during the finetuning, they could sketch the best performance under the given model capacity, and thus serve as an lower bound for WER. Note that the full model finetuning results also encompasses the results from \cite{mavandadi2023truly}, where only the 2nd pass is tunable. Fig.~\ref{fig:full} compares our LDA finetuning results with the full model finetuning results. In this figure, the yellow bars have the same meaning as Fig.~\ref{fig:over_base} denoting the WER reductions brought by LDA finetuning. The green bars represent the gaps between the LDA finetuning and the full model finetuning.

On average, the gap between ours and the full finetuning performance is less than 1\% relatively. Up to 32 languages are less than 2\% worse than the full monolingual adaptation. Even on the languages with noticeable gaps between ours and the best WERs like \textit{Czech, Hebrew}, they often have a more significant gain compared to the baseline model (yellow bars are much longer than green bars). On \textit{Czech}, while the full monolingual adaptation is 9\% better than the LDA, LDA is already 22\% better than the baseline. \textit{Hungarian} also has 35.6\% performance gain brought by our method, even though there still exists a 5\% gap to the best performance. Overall, this figure complements Fig.~\ref{fig:over_base}, illustrating that LDA adaptation can match or approach the full model finetuning in this task, and proving the value of our proposed LDA modules.

\noindent\textbf{Ablation} To evaluate the value of NST, in Fig.~\ref{fig:full}, we further compare our results with a model finetuned with the supervised data only (i.e., without NST). On average, this model is 6.4\% worse than the one with NST incorporated (blue bars). Therefore, both NST and LDA contribute to the WER reductions.
\section{Conclusion}
\label{sec:cls}

In this paper, we show promising improvements in the streaming multilingual ASR system on a tough dictation dataset. We employ the agile LDA adapters and noisy student training with the frozen foundation model, to minimize the changes required while optimizing for the performance. We test our model on a large scale with up to 39 tail languages and achieve impressive 12.2\% relative gains. In the future, we will continue to explore other adapters and improve the inference speed as well.


\footnotesize

\clearpage
\bibliographystyle{IEEEbib}
\bibliography{main_tidy}

\end{document}